\documentclass{article} %
\usepackage[final]{colm2025_conference}

\usepackage{microtype}
\usepackage{hyperref}
\usepackage{url}
\usepackage{booktabs}

\usepackage{lineno}

\definecolor{darkblue}{rgb}{0, 0, 0.5}
\hypersetup{colorlinks=true, citecolor=darkblue, linkcolor=darkblue, urlcolor=darkblue}

\usepackage{graphicx}

\usepackage{amsmath,amsfonts,bm}
\usepackage{multirow}
\usepackage[ruled]{algorithm2e}
\usepackage{textcomp}  %
\usepackage{scalerel}  %
\usepackage{booktabs}
\usepackage{pifont}
\usepackage{listings}
\usepackage{xcolor}

\def\trex{\scalerel*{\includegraphics{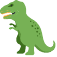}}{\textrm{\textbigcircle}}}
\newcommand{\model}{\texttt{DynaSaur\,}}
\newcommand{\cmark}{\textcolor[HTML]{2ca02c}{\ding{51}}}%
\newcommand{\xmark}{\textcolor[HTML]{d62728}{\ding{55}}}%

\title{\model\trex: Large Language Agents\\Beyond Predefined Actions}

\author{Dang Nguyen$^{1}$\thanks{Work done during internship at Adobe Research.}, Viet Dac Lai$^{2}$, Seunghyun Yoon$^{2}$, Ryan A. Rossi$^{2}$,\\
{\bf Handong Zhao$^{2}$, Ruiyi Zhang$^{2}$, Puneet Mathur$^{2}$, Nedim Lipka$^{2}$,}\\
{\bf Yu Wang$^{2}$, Trung Bui$^{2}$, Franck Dernoncourt$^{2}$, Tianyi Zhou$^{1}$}\\
$^{1}$University of Maryland, $^{2}$Adobe Research\\
\texttt{\{dangmn, tianyi\}@umd.edu} \\
}

\begin{document}

\ifcolmsubmission
\linenumbers
\fi

\maketitle

\begin{abstract}
Existing LLM agent systems typically select actions from a fixed and predefined set at every step. While this approach is effective in closed, narrowly scoped environments, it presents two major challenges for real-world, open-ended scenarios: (1) it significantly restricts the planning and acting capabilities of LLM agents, and (2) it requires substantial human effort to enumerate and implement all possible actions, which is impractical in complex environments with a vast number of potential actions. To address these limitations, we propose an LLM agent framework that can dynamically create and compose actions as needed. In this framework, the agent interacts with its environment by generating and executing programs written in a general-purpose programming language. Moreover, generated actions are accumulated over time for future reuse. Our extensive experiments across multiple benchmarks show that this framework significantly improves flexibility and outperforms prior methods that rely on a fixed action set. Notably, it enables LLM agents to adapt and recover in scenarios where predefined actions are insufficient or fail due to unforeseen edge cases.
Our code can be found in \url{https://github.com/adobe-research/dynasaur}.

\end{abstract}

\section{Introduction}
Developing autonomous agents has long been a central goal in AI research. While reinforcement learning has extensively studied this problem and has achieved significant success in specific domains \citep{DBLP:journals/nature/SilverHMGSDSAPL16, DBLP:journals/nature/SilverSSAHGHBLB17, DBLP:journals/nature/VinyalsBCMDCCPE19, DBLP:journals/nature/SchrittwieserAH20, DBLP:journals/nature/WurmanBKMS0CDE022}, it often falls short in adaptability and generalization within dynamic and uncertain environments. Given the recent advancements in Large Language Models (LLMs) \citep{DBLP:journals/corr/abs-2107-03374, DBLP:journals/corr/abs-2303-08774, DBLP:journals/corr/abs-2303-12712, DBLP:journals/corr/abs-2312-11805, DBLP:journals/corr/abs-2403-05530} with strong reasoning ability and the vast amount of world knowledge they encapsulate during pretraining, LLMs are considered promising foundations for agent policies capable of solving complex, real-world problems \citep{DBLP:journals/corr/abs-2302-04761, DBLP:journals/corr/abs-2310-05915, DBLP:conf/iclr/YaoZYDSN023, DBLP:conf/nips/DengGZCSWSS23, DBLP:conf/acl/ChenLWZLLCZ24, DBLP:conf/acl/ZengLLWLD024}. Notable initial works include Toolformer \citep{DBLP:journals/corr/abs-2302-04761}, which explores self-supervised training for LLM agents to utilize external tools, such as calculators, search engines, and translation services, thereby enhancing responses to complex question-answering tasks. ReAct \citep{DBLP:conf/iclr/YaoZYDSN023} proposes a synergistic approach by interleaving reasoning and action sequences at each step, which has become the de facto prompting framework in most LLM agent systems. Reflexion \citep{DBLP:conf/nips/ShinnCGNY23}, a follow-up work, investigates LLM agents that maintain a set of self-reflections on their past mistakes in failed trajectories; conditioning on self-reflection feedback significantly improves agent performance across various benchmarks, albeit with the trade-off of increased inference costs.

Despite these efforts, most existing LLM agent systems are studied in closed, simulated environments that accept only a finite and small set of predefined actions \citep{DBLP:conf/iclr/ZhouX0ZLSCOBF0N24, DBLP:conf/nips/Yao0YN22, DBLP:conf/nips/DengGZCSWSS23, DBLP:conf/iclr/ShridharYCBTH21, DBLP:conf/iclr/LiuGPSL18}. At every decision point, an LLM agent is constrained to select an action from this set, leading to several drawbacks. First, it restricts the agent's flexibility, preventing it from performing actions outside the predefined scope. Second, it requires significant human effort to carefully enumerate and implement all possible actions beforehand; while manageable for closed environments, this approach becomes prohibitively expensive and impractical for real-world settings. Third, in long-horizon tasks, the agent must compose sequences of primitive actions from scratch each time, limiting its ability to learn from past experiences and improve efficiency over time. To address these limitations, we propose \model, an LLM agent framework that enables the dynamic creation and composition of arbitrary actions by modeling each action as a Python function. At each step, the agent performs actions by generating Python code snippets that either define new functions, when the existing set is insufficient, or reuse existing functions from the current action set. The generated code is executed through a Python interpreter, and the resulting observations are returned to the agent. All actions generated by the agent are accumulated over time, building a library of reusable functions for future use. This approach allows the agent to extend its capabilities on the fly and compose complex actions from simpler ones, thereby enhancing its flexibility and problem-solving abilities. By leveraging the extensive ecosystem of third-party Python packages, the agent can interact with a wide range of systems and tools.

Through experiments on a diverse set of benchmarks, including GAIA \citep{DBLP:conf/iclr/MialonF0LS24}, MATH \citep{DBLP:journals/corr/abs-2103-03874}, TabMWP \citep{DBLP:conf/iclr/Lu0CWZRCK23}, AIME \citep{li2024numinamath}, and GPQA \citep{DBLP:journals/corr/abs-2311-12022}, we demonstrate that our framework enables extremely versatile LLM agents. The agent is capable of handling diverse tasks and file types without requiring human implementation of supporting functions. While the LLM agent is performant and capable on its own, extending the framework by incorporating tools developed by human experts is straightforward, simply include these tools in the agent's action set. We find that combining human-designed tools with agent-generated functions results in complementary capabilities, further enhancing the agent's performance and versatility.

\section{Problem Formulation}
We begin by formally stating our problem of interest. We model the behavior of an LLM agent as a Partially Observable Markov Decision Process defined by the tuple $(\mathcal{U}, \mathcal{A}, \mathcal{S}, \mathcal{O}, T, Z)$, where $\mathcal{U}$ is the task space; $\mathcal{A}$ is the action space, which most existing works define as a finite set of predefined actions: $\mathcal{A} = \left\{a_1, \dots, a_n\right\}$; $\mathcal{S}$ is the state space; $\mathcal{O}$ is the observation space, $T: \mathcal{S} \times \mathcal{A} \to \mathcal{P}(\mathcal{S})$ is the state transition function, mapping a state-action pair to a probability distribution over subsequent states; and $Z: \mathcal{S} \times \mathcal{A} \to \mathcal{P}(\mathcal{O})$ is the observation function, mapping a state-action pair to a probability distribution over observations. Given a task $u \in \mathcal{U}$, the agent starts in an initial state $s_0 \in \mathcal{S}$. At each time step $t$, the agent selects an action $a_t \in \mathcal{A}$ which causes the environment to transition to a new state $s_{t+1}$ according to the transition probability $T(s_t, a_t)$. The agent then receives an observation $o_{t+1} \in \mathcal{O}$ drawn from the distribution $Z(s_{t+1}, a_t)$. This process repeats until the agent reaches a terminal state $s_T$ that satisfies the original task $u$.

In this paper, we are interested in a more general setting where $\mathcal{A}$ is not fixed in advance. Specifically, we introduce a potentially infinite set $\mathcal{A}^*$ of all possible actions the agent can propose. At each time step $t$, the agent is allowed to propose any action $a_t \in \mathcal{A}^*$ to solve the task $u$. The cumulative action set at time $t$ is defined as $\mathcal{A}_t = \left\{a_1, a_2, \dots, a_t\right\}$. Each new action $a_t$ may be an entirely novel action or a composition of previously generated actions from $\mathcal{A}_{t-1}$. Consequently, the overall action space $\mathcal{A}$ evolves dynamically as the agent encounters more tasks in $\mathcal{U}$. The state transition function is accordingly redefined as $T: \mathcal{S} \times \mathcal{A}^* \to \mathcal{P}(\mathcal{S})$, and the observation function as $Z: \mathcal{S} \times \mathcal{A}^* \to \mathcal{P}(\mathcal{O})$.

\section{Methodology}

\begin{figure*}[t]
  \includegraphics[width=\textwidth]{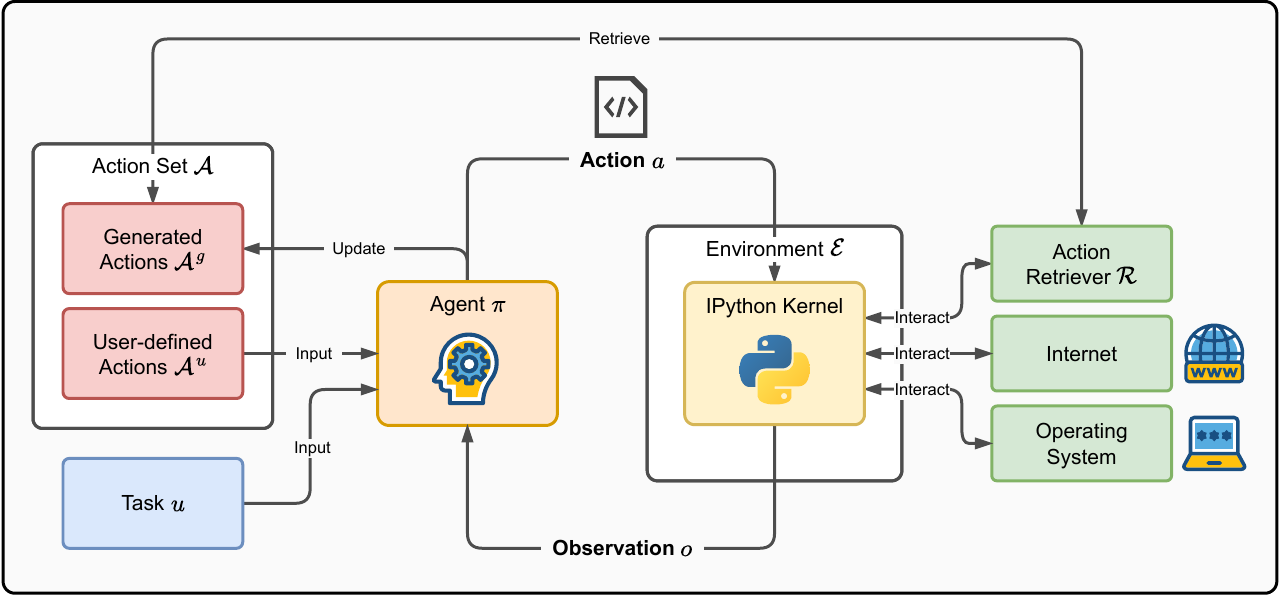}
  \caption{\textbf{Illustration of the \model agent framework}. The agent $\pi$ receives a task $t$ and optionally a set of human-designed actions $\mathcal{A}^u$. It then interacts with an environment $\mathcal{E}$ by proposing an action $a \in \mathcal{A}$, implemented as a Python function. The action is executed in an IPython kernel, which may interface with the operating system, the internet, or the action retriever as needed. The result, either the output of the function or an error message, is returned to the agent as an observation $o$. Generated actions that execute successfully are accumulated into $\mathcal{A}^g$.}
  \label{fig:pipeline}
\end{figure*}

\paragraph{Action Representation.} To design such an LLM agent system, our first challenge is to select an appropriate representation for the action space. This representation must satisfy two key criteria: (1) \textbf{Generality}: it should be expressive enough to represent actions capable of solving a wide range of tasks; and (2) \textbf{Composability}: it should naturally support the composition of actions. We argue that a general-purpose programming language meets these requirements well. We choose Python for its popularity and extensive ecosystem of libraries. This choice not only satisfies the aforementioned criteria but also facilitates seamless integration with existing tools and libraries. In our framework, each action $a \in \mathcal{A}^*$ is represented as a Python function.

\paragraph{Action Retrieval.} Including all generated actions as part of the prompt runs the risk of exceeding the context limit as the agent generates more actions. To address this issue, we decompose the action set $\mathcal{A}$ into two subsets: an optional human-designed action set $\mathcal{A}^u$ and a generated action set $\mathcal{A}^g$. Only actions in $\mathcal{A}^u$ are included in the prompt by default, allowing developers to inject domain-specific actions they consider important. To provide the agent access to actions in $\mathcal{A}^g$, we introduce an action retrieval function $R: \mathcal{Q} \times \mathbb{N} \to 2^{\mathcal{A}^g}$, where $\mathcal{Q}$ denotes the space of queries and $\mathbb{N}$ is the set of positive integers. When generating the actions, we also instruct our agent to provide a docstring describing the purpose of each action function it generates. The docstrings are then embedded to form a set of indices of the generated actions. Given a query $q \in \mathcal{Q}$ and an integer $k \in \mathbb{N}$, the function $R(q, k)$ embeds the query using the same embedding, then computes the cosine similarity between the query's embedding and each action's docstring embedding. The top-$k$ actions in $\mathcal{A}^g$ with the highest similarities are returned to the agent as part of its observations. To enable the agent to decide when to invoke action retrieval, we include the action retrieval function $R$ itself as an action in the human-designed action set $\mathcal{A}^u$. Therefore, the agent can autonomously decide to perform action retrieval by selecting $R$ during its decision-making process.

\paragraph{Action Accumulation.} Our complete pipeline is illustrated in Figure \ref{fig:pipeline}: Given a task $u \in \mathcal{U}$ and a human-designed action set $\mathcal{A}^u$ with $R \in \mathcal{A}^u$, at time step $t$, we sample a thought-action pair $(h_t, a_t) \sim \pi_\theta(a_t \mid \mathcal{A}^u, u, c_{t-1})$ following the ReAct framework \citep{DBLP:conf/iclr/YaoZYDSN023}, where $c_{t-1} = \left\{(h_1, a_1, o_1), \dots, (h_{t-1}, a_{t-1}, o_{t-1})\right\}$ represents the interaction history up to time $t-1$. The action $a_t$ is executed, and an observation $o_t$ is returned from the environment, updating the context to $c_t = c_{t-1} \cup \left\{(h_t, a_t, o_t)\right\}$. If $a_t$ contains a new function not present in $\mathcal{A}^g_{t-1}$, we update the generated action set by setting $\mathcal{A}^g_t = \mathcal{A}^g_{t-1} \cup {f(a_t)}$, where $f(a_t)$ denotes the set of functions defined in action $a_t$. Our detailed prompt can be found in Figure \ref{fig:system_prompt}. For evaluation, we employ action accumulation during training but disable it during testing. This approach ensures that performance on each test task is independent of other test tasks. %

\section{Experiments}

\begin{table*}[t]
\centering
\resizebox{\textwidth}{!}{%
\begin{tabular}{lcccccccccc}
\toprule
\multirow{2}{*}{Agent Pipeline} && \multicolumn{4}{c}{GPT-4o mini}    && \multicolumn{4}{c}{GPT-4o}      \\
\cmidrule{3-6} \cmidrule{8-11} 
                       && Level 1 & Level 2 & Level 3 & Avg.  && Level 1 & Level 2 & Level 3 & Avg.       \\
\midrule
MMAC (rep.)        && -       & -       & -       & -     && 45.16   & 20.75   & 6.12   & 25.91      \\
AutoGen Multi-Agent (rep.)        && -       & -       & -       & -     && 47.31   & 28.93   & 14.58   & 32.33      \\
HF Agent (rep.)        && -       & -       & -       & -     && 49.46   & 28.30   & 18.75   & 33.33      \\
Sibyl (rep.)           && -       & -       & -       & -     && 47.31   & 32.70   & 16.33   & 34.55      \\
Trase Agent (rep.)     && -       & -       & -       & -     && 50.54   & 33.33   & 14.29   & 35.55      \\
\midrule
No Pipeline            && 7.53    & 4.40    & 0.00    & 4.65  && 13.98   & 8.81    & 2.04    & 9.30       \\
Sibyl (repl.)          && 21.51   & 15.72   & 4.08    & 15.61 && 38.71   & 24.53   & 10.20   & 26.58      \\
HF Agent (repl.)       && 32.26   & 21.38   & \textbf{8.33}    & 22.67 && 39.78   & 27.04   & 14.58   & 29.00      \\
\model               && \textbf{45.16}   & \textbf{22.01}   & 8.16    & \textbf{26.91} && \textbf{51.61}   & \textbf{36.48}    & \textbf{18.37}   & \textbf{38.21}      \\
\bottomrule
\end{tabular}%
}
\caption{Performance comparison between various baseline methods and our proposed approach on the GAIA benchmark, evaluated under two LLM backbones: \texttt{gpt-4o-2024-08-06} and \texttt{gpt-4o-mini-2024-07-18}. ``No Pipeline'' refers to the baseline where no agent pipeline is employed, and the raw LLM is used. Results marked with (rep.) are reported results, while (repl.) indicates replicated results. Each value represents the average exact match percentage between the predicted answers and the ground truth.}
\label{tab:main_results}
\end{table*}

\begin{table*}[t]
\centering
\begin{minipage}[t]{0.52\textwidth}
\centering
\resizebox{\textwidth}{!}{%
\begin{tabular}{lcccc}
\toprule
       & No Pipeline & Sibyl System & HF Agent & DynaSaur       \\
\midrule
MATH   & 77.86       & 74.29        & 80.71    & \textbf{82.14} \\
TabMWP & 95.71       & 95.00        & 96.43    & \textbf{97.14} \\
AIME   & 13.00       & 20.00        & 20.00    & \textbf{31.71} \\
GPQA   & 48.00       & 46.00        & 38.00    & \textbf{54.00} \\
\bottomrule
\end{tabular}%
}
\caption{Performance comparison between various baseline methods on additional datasets. We utilize \texttt{gpt-4o-2024-08-06} as the LLM backbone in this experiment.}
\label{tab:more_datasets}
\end{minipage}
\hfill
\begin{minipage}[t]{0.44\textwidth}
\centering
\resizebox{\textwidth}{!}{%
\begin{tabular}{cccccccc}
\toprule
\# & AA & AI & IA & Level 1 & Level 2 & Level 3 & Avg.       \\
\midrule
1 & \xmark & \cmark & \xmark & 33.96 & 18.60 & 7.69 & 21.82       \\
2 & \cmark & \cmark & \xmark & 35.85 & 19.77 & 7.69 & 23.03       \\
3 & \xmark & \xmark & \cmark & 43.40 & 37.21 & 11.54 & 35.15      \\
4 & \xmark & \cmark & \cmark & 47.17 & 40.70 & 15.38 & 38.79      \\
5 & \cmark & \cmark & \cmark & \textbf{49.06} & \textbf{41.86} & \textbf{26.92} & \textbf{41.82}      \\
\bottomrule
\end{tabular}%
}
\caption{Ablation study on the impact of three major components: Action Accumulation (AA), Action Implementation (AI), and Initial Actions (IA).}
\label{tab:ablation_study}
\end{minipage}
\end{table*}

\subsection{Experimental Setup}
\paragraph{Benchmarks.} While numerous interactive environments exist for LLM agents, such as WebArena \citep{DBLP:conf/iclr/ZhouX0ZLSCOBF0N24}, WebShop \citep{DBLP:conf/nips/Yao0YN22}, Mind2Web \citep{DBLP:conf/nips/DengGZCSWSS23}, ALFWorld \citep{DBLP:conf/iclr/ShridharYCBTH21}, and MiniWoB++ \citep{DBLP:conf/iclr/LiuGPSL18}, they are not suitable for evaluating our proposed agent framework, as they only support a limited set of predefined actions and do not allow arbitrary action execution. We instead evaluate \model on a set of static datasets. Specifically, we consider GAIA \citep{DBLP:conf/iclr/MialonF0LS24}, a general agent benchmark covering a broad range of tasks including web browsing, file parsing and processing, symbolic reasoning, video understanding, and audio understanding. Additionally, we evaluate our agent on MATH \citep{DBLP:journals/corr/abs-2103-03874}, TabMWP \citep{DBLP:conf/iclr/Lu0CWZRCK23}, AIME \citep{li2024numinamath}, and GPQA \citep{DBLP:journals/corr/abs-2311-12022} for a more comprehensive assessment.

\paragraph{Baselines.} We compare our method with agent systems that utilize a fixed set of predefined actions, including Hugging Face Agents (HF Agent) \citep{hfagent} and Sibyl System v0.2 (Sibyl) \citep{DBLP:journals/corr/abs-2407-10718}. For the GAIA benchmark, we also include MMAC v1.1 (MMAC) \citep{DBLP:journals/corr/abs-2404-18074}, Multi-Agent Experiment v0.1 (AutoGen Multi-Agent) \citep{DBLP:journals/corr/abs-2308-08155}, and Trase Agent \citep{trase-systems-2025} for reference. Additionally, we include vanilla GPT-4o models without any agentic framework to establish a lower bound for comparison.

\paragraph{Initial Actions.} For a fair comparison with baselines, we initialize the action set with human-designed tools from Microsoft’s AutoGen \citep{DBLP:journals/corr/abs-2308-08155}, similar to HF Agent. These tools include a web browser, a file inspection tool that converts various file types into machine-readable Markdown format, and a visual question-answering tool. A detailed list of the tools and their descriptions can be found in Table \ref{tab:initial_actions}.

\paragraph{Models.} We utilize two LLM backbones for all agentic pipelines: GPT-4o (\texttt{gpt-4o-2024-08-06}) and GPT-4o mini (\texttt{gpt-4o-mini-2024-07-18}) through Azure OpenAI API. For further analyses, to save costs, we only evaluate using GPT-4o.

\paragraph{Implementation Details.} We use OpenAI's \texttt{text-embedding-3-large} as the embedding model and set the number of retrieved actions to $k=10$. We limit the maximum number of steps to 20 and set the temperature to 0.5 for all experiments. In the main experiment, we first run our agent on all examples in the validation set and accumulate the generated actions. We then freeze the action set for evaluation on the test set. 

\subsection{Main Results}
We evaluate our proposed method and compare its performance with selected baselines in Table~\ref{tab:main_results}. As shown in the table, \model significantly outperforms previous baselines for both LLM backbones across all difficulty levels of the GAIA benchmark. This demonstrates that the ability to perform arbitrary actions, combined with the capacity to accumulate actions over time, provides substantial advantages over traditional LLM agent pipelines with fixed action sets—particularly in highly complex, long-horizon tasks such as GAIA Level 2 and Level 3. In this experiment, because the exact version of GPT-4o used by HF Agent and Sibyl is unclear, we re-evaluated their pipelines under the same LLM backbones as ours to ensure a fair comparison. Their original results, as reported on the GAIA public leaderboard, are included as references. Results in Table~\ref{tab:more_datasets} further show that our method consistently outperforms the baselines on the MATH, TabMWP, AIME, and GPQA benchmarks. We include additional experiments on open-source LLMs in Appendix \ref{sec:results_on_open_source_llms}.

\subsection{Further Analysis}
In the following analysis, we use GAIA as the default dataset unless stated otherwise. Since only the GAIA validation set contains labels, we run action accumulation on 200 test examples and then evaluate on the entire validation set using the frozen action set.

\subsubsection{The Impact of Action Accumulation, Action Implementation, and Initial Actions?}
Our first analysis focuses on ablations of key components in the agent pipeline. We highlight three main components: (1) the initialization of the action set, (2) the capacity to implement arbitrary actions, and (3) the ability to accumulate actions across episodes. Notably, action accumulation depends on the agent’s ability to implement arbitrary actions, as previously generated actions must be executable.

As shown in Table \ref{tab:ablation_study}, initializing the agent with a set of human-designed actions (row 3) improves performance on GAIA by 61\% relative to the minimal baseline (row 1). This improvement is expected, as these tools are highly specialized for GAIA tasks, which often involve browsing the web or reading various file types. Adding support for arbitrary action implementation (row 4) further improves performance by 10\% (relative to row 3), and enabling action accumulation (row 5) yields an additional 7\% gain (relative to row 4). These results support the effectiveness of each component in our proposed framework.

\subsubsection{How Does Implementing Arbitrary Actions Improve Agent Performance?}
\begin{figure}[t]
  \centering
  \hspace*{-3.5em}
  \includegraphics[width=0.75\columnwidth]{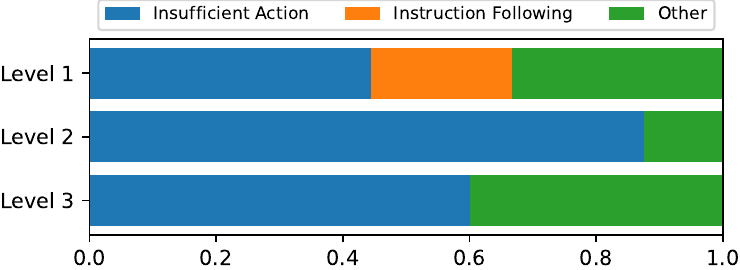}
  \caption{Distribution of error types in tasks where agent A (without action implementation) answers incorrectly, while agent B (with action implementation) answers correctly.}
  \label{fig:detailed_ai_ablation}
\end{figure}
To dive deeper into understanding the specific advantages of action implementation, we filtered out GAIA tasks that an agent without action implementation (referred to as agent A) answered incorrectly but an agent with action implementation (referred to as agent B) answered correctly. We then analyzed the reasons why agent A failed at these tasks and whether enabling action implementation in agent B helped resolve these limitations. We selected pipeline variants from row 3 in Table \ref{tab:ablation_study} as agent A and row 1 as agent B. After filtering, we obtained a set of 22 tasks. We employed OpenAI's o1 model (\texttt{o1-preview-2024-09-12}) as an evaluator. For each task, we provided o1 with the task, the correct answer, the reference trajectory from a human annotator, agent A's answer and trajectory, as well as agent B's answer and trajectory. We instructed o1 to summarize both agents' approaches with explanations for success or failure, then explain whether agent B succeeded or failed because of its ability to implement new actions. The detailed prompt is provided in Figure \ref{fig:o1_eval_prompt} in the Appendix. After o1’s evaluation, we manually analyzed the reports from o1 to further categorize agent A's errors into three types: (1) failure due to insufficient tooling, (2) failure to correctly follow instructions, and (3) failure due to other reasons.

Our findings reveal that 61.91\% of Agent A's failures were due to Reason 1, with 12 cases where the agent lacked the necessary tools to solve the task, and 1 case where a human-designed tool failed to return relevant information. In 9.52\% of the cases, agent A failed due to reason 2 (e.g., returning an answer with an incorrect unit). The remaining 28.57\% of the failures were caused by other unrelated factors, such as the inability to find relevant information online or getting stuck without making progress. A more detailed breakdown of the error distribution for each level is shown in Figure \ref{fig:detailed_ai_ablation}. In all type-1 errors, agent B was able to complete the task by implementing custom actions. This result demonstrates that our framework significantly improves the agent’s flexibility in problem solving.

\subsubsection{How Useful Are Generated Actions on Unseen Tasks?}

\begin{figure}[t]
  \centering
  \begin{minipage}[t]{0.48\linewidth}
    \centering
    \includegraphics[width=\linewidth]{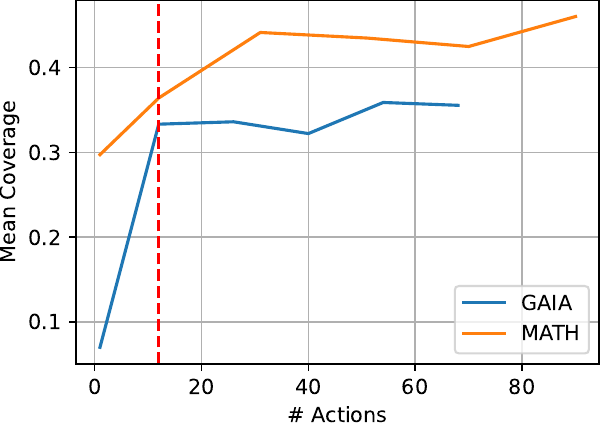}
    \caption{Mean coverage over the validation set as the number of actions increases. The red dashed line marks the point where human-designed actions are added.}
    \label{fig:coverage}
  \end{minipage}
  \hfill
  \begin{minipage}[t]{0.48\linewidth}
    \centering
    \includegraphics[width=\linewidth]{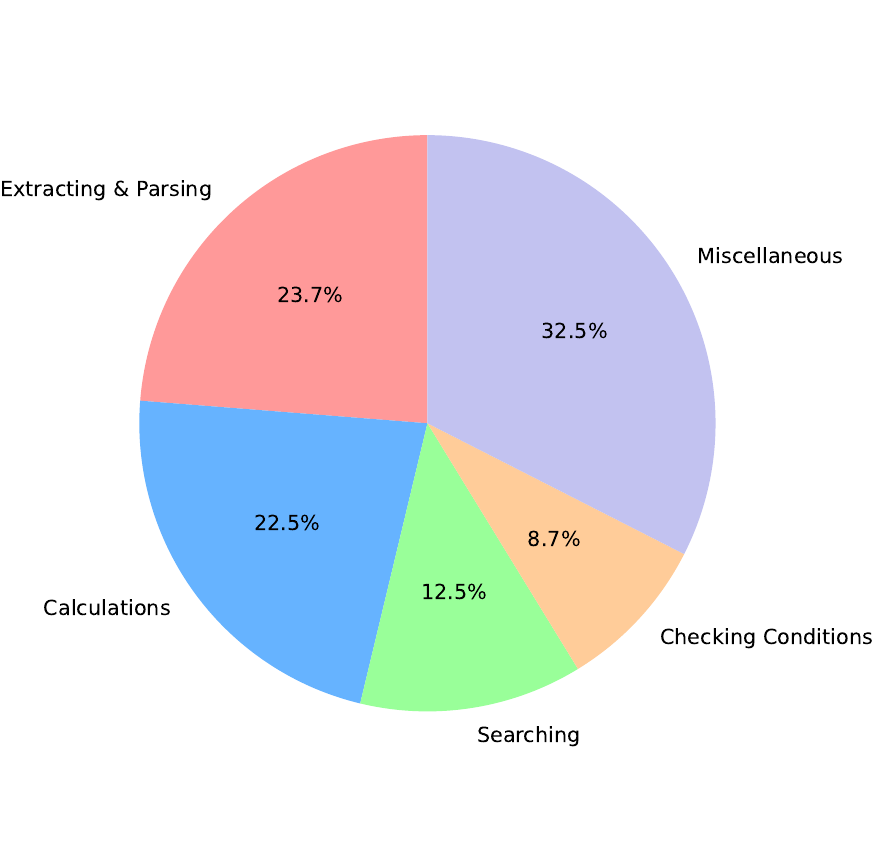}
    \caption{Categories of actions accumulated during evaluation on GAIA validation set.}
    \label{fig:action_categories}
  \end{minipage}
\end{figure}

\begin{figure*}[t]
  \includegraphics[width=\textwidth]{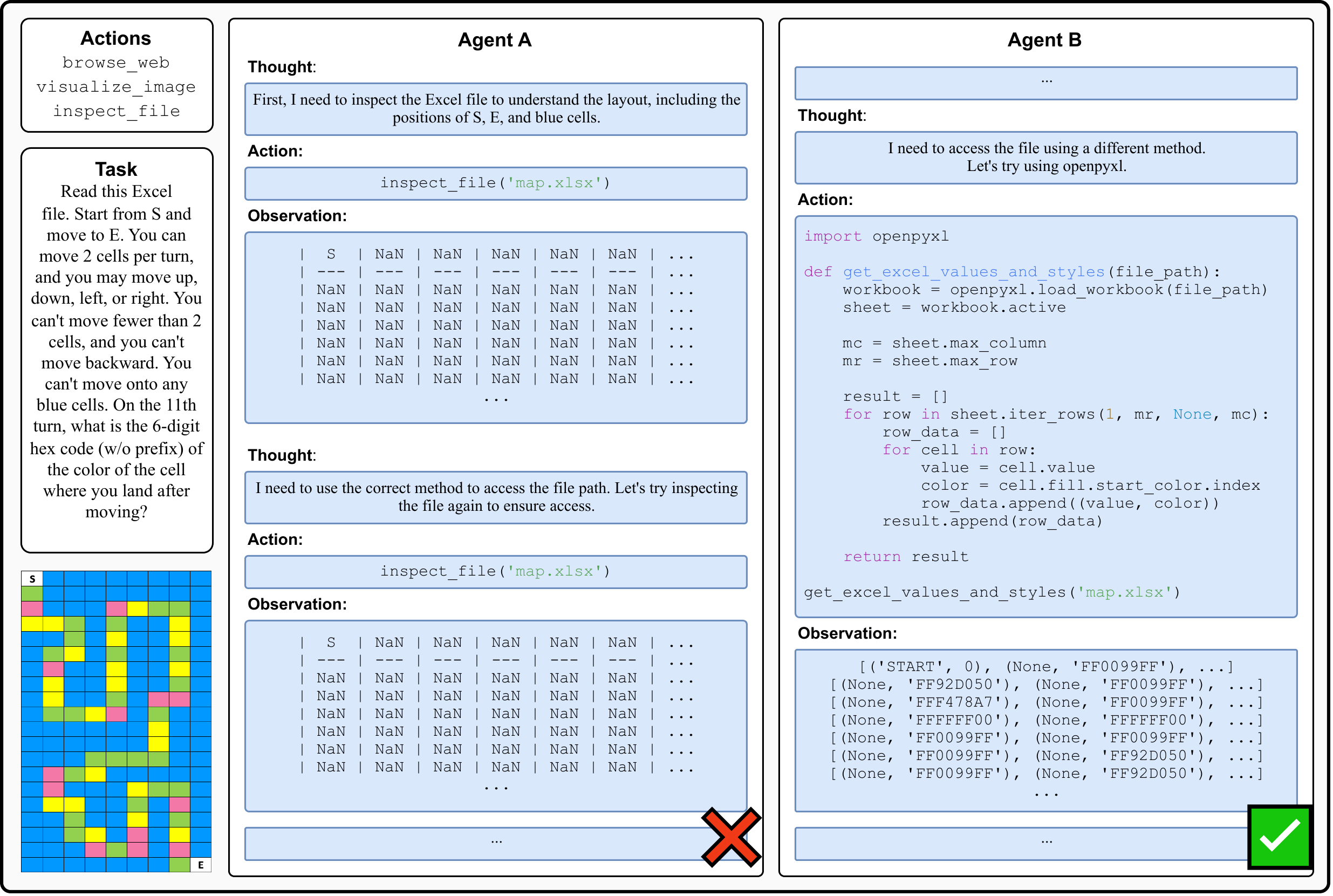}
  \caption{A case study demonstrates the difference in problem-solving flexibility between Agent A (a variant of \model without action implementation) and Agent B (the proposed agent framework). Both agents begin with the same initial step, but only Agent B, equipped with the ability to implement its own actions, successfully completes the task. Due to space constraints, the first step taken by Agent B is not shown.}
  \label{fig:case_study_1}
\end{figure*}
In this experiment, we evaluate how the utility of a generated action set improves as the agent accumulates more actions over time. To quantify this utility, we propose an \emph{Action Coverage} metric, which measures how often an agent relies on the actions in a given set $\mathcal{A}$ to solve an unseen task $u$. Formally, given a task $u$, a ground-truth answer $y$, and an action set $\mathcal{A}$, we sample a trajectory $\tau = \left\{ \left( a_1, o_1 \right), \dots, \left( a_T, o_T \right) \right\}$ from the policy $\pi_\theta(\cdot \mid \mathcal{A}, u)$, where $a_i$ and $o_i$ denote the action and observation at step $i$ (we omit the intermediate thought sequence $h_i$ for brevity). We consider the task successfully solved if the final observation $o_T = y$. At each step, the agent either reuses an action from $\mathcal{A}$ or generates a new one not in $\mathcal{A}$. If the agent must generate a new action to complete the task, we consider the action set $\mathcal{A}$ insufficient for that task. We define the \emph{coverage} of $\mathcal{A}$ over $u$ as the proportion of steps where the agent uses actions from $\mathcal{A}$, conditioned on successful task completion:
\begin{equation}
    C(\mathcal{A}, u) \overset{\underset{\mathrm{def}}{}}{=}\underset{\tau\sim \pi_\theta(\cdot|\mathcal{A}, u)}{\mathbb{E}}\biggl[1 - \dfrac{1}{|\tau|}\mathbf{1}[o_T = y]\cdot\Big|\left\{(a_i, o_i)\in\tau:a_i \notin \mathcal{A}\right\}\Big|\biggr]
\end{equation}
However, computing this metric exactly is prohibitively expensive. In practice, we approximate it by sampling a single trajectory for each task. We report the average Action Coverage over the test sets of GAIA and MATH, as visualized in Figure~\ref{fig:coverage}.

Initially, when $|\mathcal{A}| = 1$ (consisting only of the \texttt{submit\_final\_answer} action), coverage is low for both datasets, as expected. However, it is not zero, since a small number of tasks in both GAIA and MATH can be solved by directly submitting the final answer at the first step. When human-designed tools are added (marked by the red dashed line), Action Coverage increases substantially for GAIA, but less so for MATH. This is because the tools are well-aligned with the steps required in GAIA tasks, such as web browsing and file reading, but are less applicable to mathematical problem solving. In the subsequent steps, as the agent accumulates generated actions, coverage increases for both datasets, with a stronger rise observed in MATH. Our analysis suggests this is due to the domain mismatch: on GAIA, the agent continues to rely heavily on the human-designed tools; whereas on MATH, where those tools are less useful, the agent increasingly relies on its own generated~tools.

\subsection{Statistics of the Generated Actions}
We present summary statistics and analyses of the actions generated during evaluation on GAIA tasks. This section includes a breakdown of action types, categorization by functionality, and a complexity analysis of the generated code. Additionally, we provide concrete examples of both successful and failed actions in the Appendix.

\subsubsection{Action Statistics and Categories}
A total of 174 actions were generated, comprising 80 actions accumulated during training (from 165 examples) and 94 newly generated during testing (from 300 examples). We manually inspected each action and grouped them into categories based on their functionality. The distribution is shown in Figure~\ref{fig:action_categories}. Specifically, 23.75\% of the actions are dedicated to extracting and parsing information from data or files, 22.5\% perform calculations, 12.5\% involve searching operations, 8.75\% check conditions (e.g., evaluating whether a statement is true or false), and the remaining 32.5\% fall into a miscellaneous category, including tasks such as file conversion and counting.

\subsubsection{Complexity Analysis}
To assess the complexity of the generated actions, we use the cyclomatic complexity metric \citep{1702388}, which quantifies the number of linearly independent paths in a program’s control flow. A cyclomatic complexity below 10 is generally considered a threshold for maintainable code, while higher values may indicate more intricate and error-prone logic. Based on this metric, the generated actions exhibit an average complexity of 3.06, slightly lower than the average of 3.72 observed in human-authored actions.

\subsection{Case Studies}
We present a real case study comparing how an agent without action implementation (denoted as agent A) and an agent with action implementation (denoted as agent B) approach the same problem. In this example, the task requires the agents to load an Excel file containing a map, as shown in the lower left corner of Figure \ref{fig:case_study_1}. The agent must then navigate through the map according to the task's movement rules and, after the 11th turn, return the color of the current cell. The provided action set is similar to previous experiments. In this scenario, the \texttt{inspect\_file} tool, developed by Microsoft's AutoGen \citep{DBLP:journals/corr/abs-2308-08155}, assists an agent by reading diverse file types and returning the file content in Markdown format. However, when reading Excel files, the tool does not account for formatting properties such as cell color, leading to incomplete information being returned and preventing agent A from solving the task. Since agent A lacks other tools, it repeatedly attempts to invoke the \texttt{inspect\_file} tool until the maximum iteration limit is reached. On the other hand, agent B also initially tries to invoke the same tool but recovers from the error by using a different approach to read the Excel file content through openpyxl. In subsequent steps, agent B implements the solution for map navigation as a function and successfully completes the task (we omit the full steps due to space constraints). We include additional case studies on the benefits of dynamic action creation in Appendix \ref{sec:more_case_studies}.

\section{Related Work}
\subsection{LLM Agents}
Most current methods that utilize LLMs for agent tasks involve prompting techniques \citep{yao2023react, liang2023code, gao2023pal, kim2023language, DBLP:journals/corr/abs-2404-18074}, supervised fine-tuning \citep{schick2023toolformer, zeng2023agenttuning, chen2024agentflan, zhang2024agentohana, chen2023fireact, DBLP:journals/corr/abs-2402-11651}, or reinforcement learning (RL) algorithms for self-exploration \citep{zhou2024archer, song2024trial, yang2024react, aksitov2023rest, christianos2023panguagent, abdulhai2023lmrl, gulcehre2023reinforced, song-etal-2024-trial}. However, these approaches mainly study agents under the assumption that the set of actions is fixed and provided by the environment. Furthermore, most existing work uses text \citep{schick2023toolformer} or JSON \citep{qin2023toolllm} as the representation of actions, which significantly lacks the two criteria mentioned earlier: generality and composability. In contrast, \model can utilize available actions or create new ones if necessary, using code as a unified representation. In principle, acting with code enables agents to solve any Turing-complete problem.

\subsection{LLM Agents for Code Generation}
Although using LLMs to generate code is not new, these approaches have a long history dating back to the early stages of LLM development \citep{chen2021evaluating, austin2021program, hendrycks2021measuring}. However, this line of research has primarily focused on using LLMs as software engineering assistants for tasks like code completion or program synthesis \citep{austin2021program, zhang2024codeagent}. In our work, we utilize programming languages as a tool to solve generalist AI agent tasks in the GAIA benchmark, which require multistep execution in partially observable and stochastic environments.

\subsection{LLM Agents for Tool Creation}
There have been a few attempts to explore LLMs' ability to create their own tools, though these efforts have largely been limited to solving simple problems \citep{Cai2023LargeLM, Qian2023CREATORTC, wang2023voyager, Yuan2023CRAFTCL}. For example, \citep{Cai2023LargeLM} examines LLMs generating code snippets to tackle basic tasks such as word sorting or simple logical deduction. Their approach involves sampling three input-output pairs of a specific task type, using the LLM to generate a function to solve the problem, validating it with three additional pairs from the validation set, and then evaluating the solution on all test instances from the same task type. This setup simplifies the problem as the task type remains consistent during both training and testing. Similarly, \citep{Qian2023CREATORTC} and \citep{Yuan2023CRAFTCL} explore tool creation, but restrict their focus to math problems, with \citep{Yuan2023CRAFTCL} also introducing VQA benchmarks. These tasks are typically solvable in a single step and do not require interaction with an external environment. We are the first to study generalist LLM agents that implement and accumulate actions within the real-world agent benchmark GAIA.

\section{Conclusion}
We propose a novel LLM agent framework that leverages Python as a universal representation for actions. By using a general-purpose programming language, our framework enables the implementation of arbitrary actions as well as compositions of existing ones—effectively addressing the limitations of prior agent systems that rely on a fixed, predefined action set. This design not only enhances expressiveness but also allows the agent to perform more complex, context-specific reasoning and decision-making. Moreover, our framework supports unsupervised accumulation of new actions over time, making it suitable for both online and offline deployment scenarios. This adaptability enables the agent to continually expand its capabilities without manual intervention or retraining. We believe that such flexibility is key to achieving strong generalization across diverse tasks and environments. To validate our approach, we conduct extensive experiments across a variety of challenging benchmarks, including GAIA, MATH, TabMWP, AIME, and GPQA. Results consistently demonstrate the effectiveness of our framework. In addition, our analysis reveals that the agent is capable of autonomously recovering from tool failures caused by unforeseen edge cases—highlighting its robustness in real-world settings.

\section*{Ethics Statement}
As a proof of concept, our framework allows agents to generate and execute arbitrary Python code. While this approach is not advisable for real-world deployment due to potential security risks, we acknowledge that various safeguards can be implemented to mitigate these concerns. For example, one can apply a safety filter or a world-model-based formal verifier to each action during creation or prior to execution. Furthermore, the agent should be deployed in an isolated environment with restricted inbound and outbound traffic. Limiting file system permissions—such as enforcing read-only access or encouraging minimal edits instead of overwriting files—can further reduce the risk of unintended or harmful behavior. Restricting the agent’s permissions also helps prevent the execution of malicious scripts.

\bibliography{colm2025_conference}
\bibliographystyle{colm2025_conference}

\appendix
\newpage

\section{Additional Experiments on Open-source Large Language Models}\label{sec:results_on_open_source_llms}

\begin{table*}[t]
\centering
\resizebox{\textwidth}{!}{%
\begin{tabular}{lcccccccccc}
\toprule
\multirow{2}{*}{Agent Pipeline} && \multicolumn{4}{c}{Qwen2.5-32B-Instruct}    && \multicolumn{4}{c}{Qwen2.5-Coder-32B-Instruct}      \\
\cmidrule{3-6} \cmidrule{8-11} 
                       && Level 1 & Level 2 & Level 3 & Avg.  && Level 1 & Level 2 & Level 3 & Avg.       \\
\midrule
No Pipeline            && 5.66    & 5.81    & 0.00    & 4.85  && 13.21   & 1.16   & 0.00   & 4.85       \\
Sibyl                  && 24.53   & 10.47   & 0.00    & 13.33 && 20.75   & 8.14   & 0.00   & 10.91      \\
HF Agent               && 26.42   & 11.63   & \textbf{3.85}    & 15.15 && 13.21   & 10.47  & 3.85   & 10.30      \\
\model                 && \textbf{35.85}   & \textbf{30.23}   & \textbf{3.85} & \textbf{27.88} && \textbf{35.85}   & \textbf{20.93}    & \textbf{11.54}   & \textbf{24.24}      \\
\bottomrule
\end{tabular}%
}
\caption{Performance comparison between various baseline methods and our proposed approach on the GAIA benchmark, evaluated under two LLM backbones: \texttt{Qwen2.5-32B-Instruct} and \texttt{Qwen2.5-Coder-32B-Instruct}. ``No Pipeline'' refers to the baseline where no agent pipeline is employed, and the raw LLM is used. Each value represents the average exact match percentage between the predicted answers and the ground truth.}
\label{tab:results_on_open_source_llms}
\end{table*}

To evaluate whether our method generalizes beyond proprietary models and remains effective when paired with models with weaker code generation abilities, we tested two open-source models: \textbf{Qwen2.5-Coder-32B-Instruct}, a code-specialized, instruction-tuned model, and \textbf{Qwen2.5-32B-Instruct}, a general-purpose, instruction-tuned model with weaker code generation capabilities. The results, shown in Table~\ref{tab:results_on_open_source_llms}, highlight that DynaSaur works effectively with open-source models, not just proprietary LLMs. Even with a general-purpose model like Qwen2.5-32B-Instruct, which has weaker code generation abilities, DynaSaur significantly outperforms other pipelines. Surprisingly, Qwen2.5-32B-Instruct even slightly outperforms its code-specialized variant. We hypothesize that its stronger commonsense reasoning and decision-making capabilities may compensate for its weaker coding skills, ultimately resulting in better overall performance.

\section{Implementation Details}\label{sec:appendix}
\subsection{Initial Actions}
\begin{table*}[t]
\small
\centering
\resizebox{\textwidth}{!}{%
\begin{tabular}{c p{0.25\linewidth} p{0.75\linewidth}}
\toprule
\# & Action Header                                                      & Description    \\                                                                                   
\midrule
1  & \texttt{submit\_final\_answer}      & Submits the final answer to the given problem.                                                                                    \\
2  & \texttt{get\_relevant\_actions}     & Retrieve $k$ most relevent generated actions given a query.                                                                       \\
3  & \texttt{informational\_web\_search} & Perform an informational web search query then return the search results.                                                         \\
4  & \texttt{navigational\_web\_search}  & Perform a navigational web search query then immediately navigate to the top result.                                              \\
5  & \texttt{visit\_page}                & Visit a webpage at a given URL and return its text.                                                                               \\
6  & \texttt{download\_file}             & Download a file at a given URL.                                                                                                   \\
7  & \texttt{page\_up}                   & Scroll the viewport up in the current webpage and return the new viewport content.                                \\
8  & \texttt{page\_down}                 & Scroll the viewport down in the current webpage and return the new viewport content.                              \\
9  & \texttt{find\_on\_page\_ctrl\_f}    & Scroll the viewport to the first occurrence of the search string.                                                                 \\
10 & \texttt{find\_next}                 & Scroll the viewport to next occurrence of the search string.                                                                      \\
11 & \texttt{find\_archived\_url}        & Given a url, searches the Wayback Machine and returns the archived version of the url that's closest in time to the desired date. \\
12 & \texttt{visualizer}                 & Answer question about a given image.                                                                                              \\
13 & \texttt{inspect\_file\_as\_text}    & Read a file and return its content as Markdown text.           \\         
\bottomrule
\end{tabular}%
}
\caption{List of initial actions used in this project.}
\label{tab:initial_actions}
\end{table*}

We present the list of initial actions used in this project, along with their descriptions, in Table \ref{tab:initial_actions}. Actions 3 to 13 are adopted from Microsoft's AutoGen \citep{DBLP:journals/corr/abs-2308-08155}.

\subsection{Prompt For Qualitative Analysis}\label{sec:action_impl_eval_prompt}
\begin{figure*}[t]
  \includegraphics[width=\textwidth]{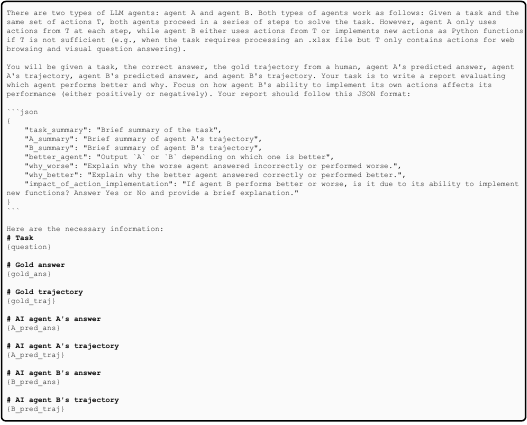}
  \caption{Prompt for OpenAI's o1 to perform qualitative evaluation.}
  \label{fig:o1_eval_prompt}
\end{figure*}

The prompt for qualitative analysis with OpenAI's \texttt{o1-preview} model is shown in Figure \ref{fig:o1_eval_prompt}.

\subsection{\model's System Prompt}\label{sec:system_prompt}
\begin{figure*}[t]
  \includegraphics[width=\textwidth]{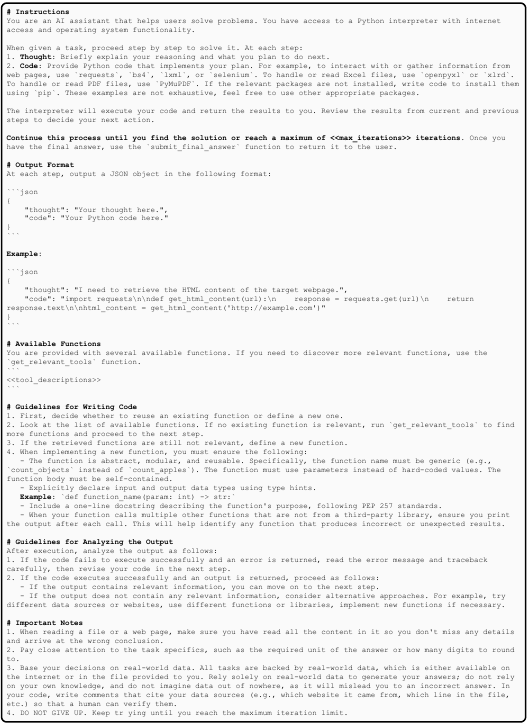}
  \caption{The system prompt of our \model agent framework.}
  \label{fig:system_prompt}
\end{figure*}

The system prompt used for \model is shown in Figure \ref{fig:system_prompt}.

\section{Additional Case Studies}\label{sec:more_case_studies}
\begin{figure*}[t]
  \includegraphics[width=\textwidth]{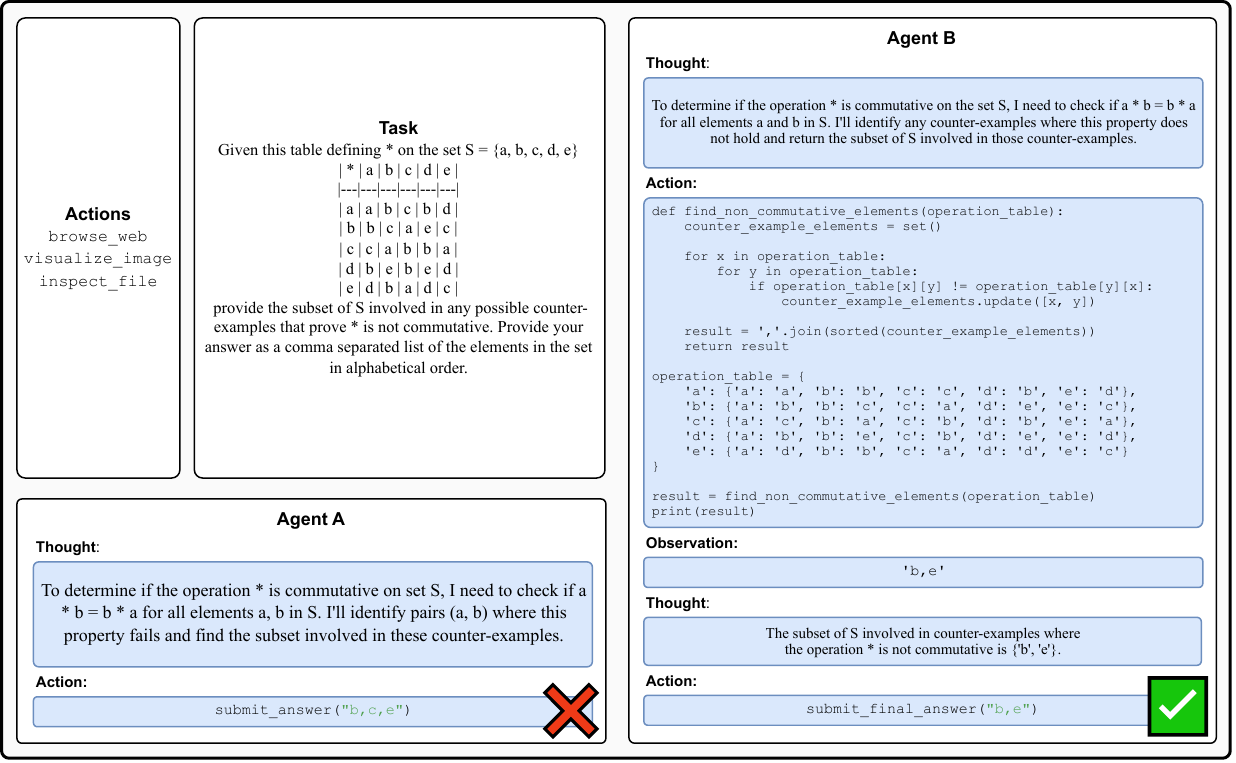}
  \caption{A case study demonstrates the difference in problem-solving flexibility between Agent A (a variant of \model without action implementation) and Agent B (the proposed agent framework).}
  \label{fig:case_study_2}
\end{figure*}

We present another comparative case study of two agents: one without action implementation (referred to as agent A) and one with action implementation (referred to as agent B), illustrated in Figure~\ref{fig:case_study_2}. In this scenario, both agents are provided with a binary operator $*$ defined by a table and tasked with finding a counterexample to demonstrate that $*$ is not commutative. Successfully solving this task requires symbolic reasoning abilities. Agent A, lacking the necessary actions to address this task thoroughly, attempts reasoning within its Thought sequence but ultimately submits an incorrect answer. In contrast, agent B dynamically generates a specialized function to tackle the question. This action is general enough to solve other instances of the original problem as well. This example further highlights the advantage of equipping agents with the ability to dynamically generate and execute actions through code to tackle a range of problems.

\section{Examples of Generated Actions.}
\begin{figure}[t]
\centering
\begin{minipage}{0.95\textwidth}
\begin{lstlisting}
import fitz

def extract_text_from_pdf(file_path: str) -> str:
    """Extract text from a PDF file."""
    text = ''
    with fitz.open(file_path) as pdf:
        for page in pdf:
            text += page.get_text()
    return text

from openpyxl import load_workbook
 
def inspect_excel_file(file_path: str):
    """Inspect data from an Excel file."""
    workbook = load_workbook(filename=file_path)
    sheet = workbook.active
    data = []
    for row in sheet.iter_rows(values_only=True):
        data.append(row)
    return data
\end{lstlisting}
\end{minipage}
\caption{Successful examples of generated actions in GAIA.}
\label{fig:successful_code}
\end{figure}

\begin{figure}[t]
\centering
\begin{minipage}{\textwidth}
\begin{lstlisting}
def calculate_food_sales(sheet) -> float:
    """Calculate the total sales from food items in the given Excel sheet."""
    total_sales = 0.0
    for row in sheet.iter_rows(min_row=2, values_only=True):
        total_sales += sum(row[1:6])
    return total_sales

def count_crustacean_mentions(slide_text: str) -> int:
    """Count slides mentioning crustaceans in the provided slide text."""
    crustaceans = ['crayfish', 'isopods', 'Yeti crab', 'Spider crab']
    count = 0
    for crustacean in crustaceans:
        if crustacean in slide_text:
            count += 1
    return count
\end{lstlisting}
\end{minipage}
\label{fig:failed_code}
\caption{Failed examples of generated actions in GAIA.}
\end{figure}

We include examples of both successful and failed generated actions in Figures \ref{fig:successful_code} and \hyperlink{10}{10}. A generation is considered successful when the action is reasonably generalizable and applicable across various contexts. Conversely, an action is considered a failed generation if it contains hard-coded values or is too context-specific to be reused in different tasks.

\end{document}